\documentclass{article}
\usepackage{spconf,amsmath,graphicx,hyperref}
\usepackage{url}
\usepackage{amsmath,amssymb,amsfonts}
\usepackage{algorithmic}
\usepackage{algorithm}
\usepackage{multirow}
\usepackage{booktabs} 
\usepackage{textcomp}
\usepackage{xcolor}
\newtheorem{remark}{Remark}
\usepackage{graphicx}
\usepackage{float} 
\usepackage{subfigure}
\usepackage{multicol}
\usepackage{setspace}

\begin{document}
\title{CADM: Cluster-customized Adaptive Distance Metric \\ for Categorical Data Clustering}
\name{Taixi Chen$^{1}$ \qquad Yiu-ming Cheung$^{2}$$^{\dagger}$\thanks{$^{\dagger}$Corresponding author: ymc@comp.hkbu.edu.hk} \qquad Yiqun Zhang$^{2}$}
  
\address{$^{1}$ School of Computing, Binghamton University, Binghamton, NY, USA \\ $^{2}$ Department of Computer Science, Hong Kong Baptist University, Hong Kong SAR, China}
%
\maketitle
\begin{abstract}
An appropriate distance metric is crucial for categorical data clustering, as the distance between categorical data cannot be directly calculated. However, the distances between attribute values usually vary in different clusters induced by their different distributions, which has not been taken into account, thus leading to unreasonable distance measurement. Therefore, we propose a cluster-customized distance metric for categorical data clustering, which can competitively update distances based on different distributions of attributes in each cluster. In addition, we extend the proposed distance metric to the mixed data that contains both numerical and categorical attributes. Experiments demonstrate the efficacy of the proposed method, i.e., achieving an average ranking of around first in fourteen datasets. The source code is available at \href{https://github.com/Taixi-CHEN/CADM}{https://github.com/Taixi-CHEN/CADM} 
\end{abstract}

\section{Introduction}
Cluster analysis of categorical data composed of nominal and ordinal attributes are common in many fields, such as medical analysis, customer questionnaires, and so on \cite{book, cc, cc1, cc2}. Nevertheless, due to the difficulty of measuring the difference between categorical attributes, the core problem of categorical data clustering relies on discovering and defining a proper distance metric for effective measurement. The existing distance metrics have been explored along two main branches: 1) directly calculate the distance of categorical data based on the defined encoding methods \cite{hdm, early_abdm, sbc}, and 2) indirectly estimate the distance between different attributes based on the frequency or distribution in context \cite{oldest_jdm, Condist}. However, most of them neglect the heterogeneity between ordinal and nominal attributes in categorical data.

\begin{table*}[t]
    \centering
    \begin{minipage}{0.49\textwidth} 
    \centering
        \resizebox{.7\linewidth}{!}{
        \begin{tabular}{c|c c c }
        \toprule
          ID & finance & social & health \\
          \midrule
            People\_1 & convenient & nonprob & recommended  \\
            People\_2 & convenient & nonprob & priority  \\
            People\_3 & convenient & nonprob & not\_recom  \\
            People\_4 & convenient & slightly\_prob & recommended  \\
            People\_5 & convenient & slightly\_prob & priority  \\
            People\_6 & convenient & problematic & recommended  \\
            People\_7 & convenient & problematic & priority  \\
            People\_8 & convenient & problematic & not\_recom  \\
            People\_9 & inconv & nonprob & recommended  \\
            People\_10 & inconv & nonprob & priority  \\
            \bottomrule
        \end{tabular}}

        \vspace{.4cm}
        \text{(I)}
    
    \end{minipage}%
    \begin{minipage}{0.5\textwidth} 
        \centering
        \includegraphics[width=.9\linewidth]{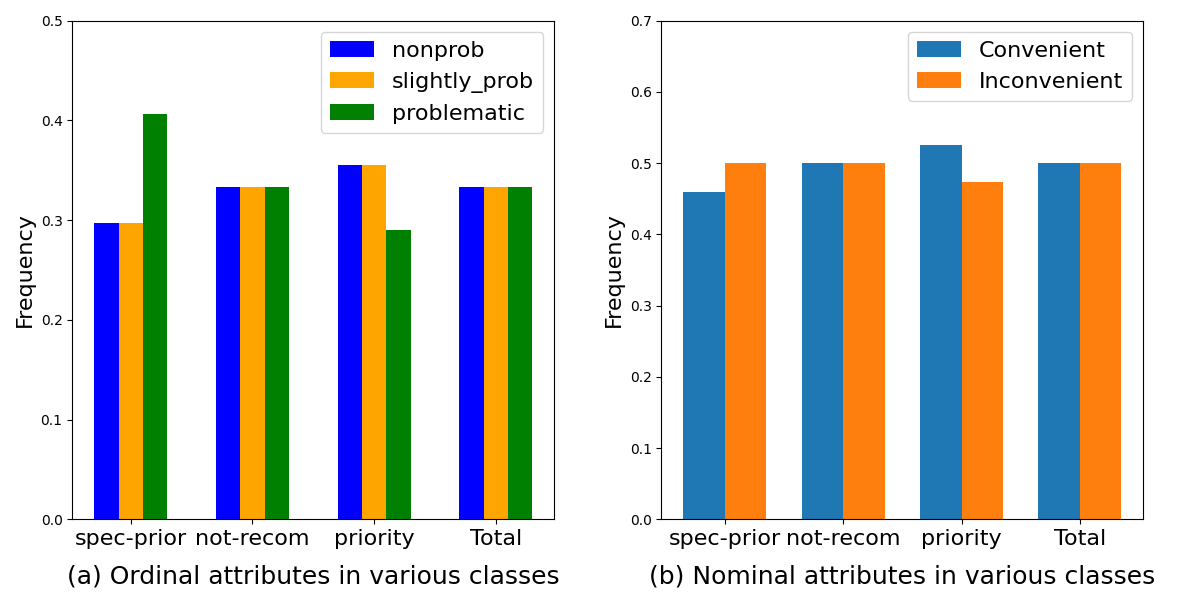} 
        
    \text{(II)}
    \end{minipage}
    
    \caption{ (I) The examples in the Nursery categorical dataset. The finance attribute is nominal, while others are ordinal. (II) The distance and distribution of both ordinal and nominal attributes are different in each cluster.}
    \label{fig:distribution}
\end{table*}

Recently, the order information of ordinal data has received increasing attention \cite{old_order, udm} because order reflects the intrinsic difference between ordinal attribute values. For instance, in the Nursery categorical dataset (Table~\ref{fig:distribution} (I)), its ordinal attribute $social$ contains three ordinal values: $nonprob$, $slightly\_prob$, and $problematic$. The distance between $nonprob$ and $problematic$ should not only consider their difference, as their semantic concepts are not isolated and independent, but related to their medium value $slightly\_prob$ as well \cite{ordertry}. 

However, existing methods consider the order information, the intrinsic distance between ordinal attribute values, to be the same in the entire dataset, ignoring the heterogeneity of different clusters. It is not reasonable in many cases, for example, in Table~\ref{fig:distribution} (II), the distance between $nonprob$ and $problematic$ in the class spec-prior and priority is larger than that in class not-recom, based on their frequency in context, because their importance is different in different classes. Moreover, Table~\ref{fig:distribution} (II) also shows that the distance of nominal attribute values varies in three classes induced by the distinction of their frequency distributions in different classes. Thus, it can be observed from the two sub-figures that the commonly used total context frequency distribution cannot reflect this distribution difference for both nominal and ordinal attribute values between different clusters, which restricts the performance of the distance metric and clustering.

To tackle these challenges, this paper proposes a novel distance metric called Cluster-customized Adaptive Distance Metric (CADM). This is a unified distance metric for both ordinal and nominal data. It defines the attribute value distances between objects and different cluster centers as Cluster-customized attribute Value Distance (CVD), depending on Cluster-customized Value Importance (CVI), adaptively changing in different clusters during iterations. Specifically, CVI is the importance of one attribute value in different clusters, which is determined by both this attribute value's count and the maximum count of all attribute values in this attribute. Based on CVD, the data with high cluster importance attribute values will be pulled closer to the cluster center, as it represents this cluster. Otherwise, it will be pulled away from the cluster center. The intuition behind this idea is to reasonably leverage information guidance from different clusters to improve distance measurement. Based on the observation of Table~\ref{fig:distribution}, it is necessary to design more refined distance measurements for attribute values.

Furthermore, a Cluster-customized Attribute Importance (CAI) is defined to weigh attribute contributions in forming distances, which regards the consistency of possible attribute values in one attribute category. This mechanism is applicable in attribute-independent cases as it depends on the self-importance. In addition, we extend CADM to mixed data with heterogeneous attributes in the experiment. In summary, this paper makes the following contributions:
\begin{itemize}
\item A unified distance metric CADM is proposed for nominal and ordinal data that considers the adaptive cluster-customized distance measurement, addressing the problem of distance difference in various clusters.

\item Based on the CVI, the CVD is defined to dynamically measure the attribute value distance between categorical data and the cluster center. It can provide personalized measurements for each cluster, reducing bias during the clustering process.

\item To weigh the attribute contributions in forming distances, this paper defines the CAI, which can achieve minute adjustments to CVD, making distance measurement more reasonable and accurate.
\end{itemize}

\section{Proposed method}
\label{sec: Proposed method}
In this section, we first formulate the problem. Then, we introduce our proposed distance metrics CADM and provide the algorithm analysis of it.

\subsection{Problem Formulation}

Assuming a dataset $S$ can be rewritten as $S = <X, A, O>$. The data object sets $X =[x_0, x_1, ..., x_{n-1}] $ with $n$ objects. And for each sample $x_i = [x_i^0, x_i^1, ..., x_i^{d-1}]$ because it has the $d$ attributes. Moreover, as the attribute $A = [A^0, A^1, ..., A^{d-1}]$, for each attribute, it has $n$ values so that $A^r = [A^r_0, A^r_1, ..., A^r_{n-1}]$. Besides, as each attribute $A^r$ must have limited possible values $v^r$, thus, the unique set $O^r$ for different attribute $A^r$ can be written as $O^r = [o^r_0, o^r_1, ..., o^r_{v^r-1}]$, which is ascending order. Besides, data should be placed as $A$ = $A^{num}$ + $A^{nom}$ + $A^{ord}$, while the numerical data $A^{num}$ is optional. It is worth noting that mixed and categorical data clustering normally adopts the k-prototypes clustering algorithm \cite{KMD, udm, HARR}, which only considers the distance between categorical data and cluster centers. Each cluster is described by a center $c_l = [c_l^0, c_l^1, ..., c_l^{d-1}]$ from $C = [c_0, c_1, ..., c_{k-1}]$. It aims to assign $n$ data objects in $X$ to $k$ proper clusters, which can be formulated as minimizing:
\begin{equation}
    J = \sum_{i=0}^{n-1}\sum_{l=0}^{k-1} d(x_i, c_l)
\label{obj}    
\end{equation}
where $c_l$ is one specific cluster center, and the value of $c_l^r$ is the most frequent possible value from $A^r$ in $l_{th}$ cluster. The dissimilarity between an object and the cluster center can be rewritten as \begin{equation}
    d(x_i, c_l) = \sum_{r=0}^{d-1} d_m(x^r_i, c^r_l) + d_I(A^r),
\label{eq: logic}
\end{equation} where the $d_m(.)$ is the distance between the categorical attribute values, and $d_I(.)$ measures the importance of the categorical attribute $A^r$. 
\subsection{Cluster-customized Adaptive Distance Metric}
CADM is proposed to adaptively measure the cluster-personalized distance between categorical data. Thus, we define the distances of different categorical attribute values, such as $x^r_i$ and $c^r_l$, as computed by:
\begin{equation}
d_m(x^r_i, c^r_l)=\left\{
\begin{array}{l l}
\sum_{j = \min(\alpha(x^r_i, c^r_l))}^{\max(\alpha(x^r_i, c^r_l))}d_a^l(o_j^r, o_p^r), & A^r \in A^{ord} \\[1ex]
d_a^l(o_t^r, o_p^r), & A^r \in A^{nom}
\end{array} \right.
\label{eq:intra1}
\end{equation}
where $x_i^r$ and $c^r_l$ are denoted as $o_t^r$ and $o_p^r$ in $O^r$ perspective. The $o_j^r$ represents the intermediate attribute value between $x^r_i$ and $c^r_l$, including $x_i^r$ as well. If attribute $A^r$ is an ordinal attribute, CADM uses order information from the intermediate attribute value following existing works \cite{udm, learn} to enhance measurement. The $d_{a}^l(.)$ is the CVD designed for measuring the distance of attribute values. The notation $l$ means distance measure in the $l_{th}$ cluster. The $\alpha(x^r_i, c^r_l)$ fetches the order number of $x^r_i$ and $c^r_l$ from $O^r$. In addition, the $d_m(x^r_i, c^r_l)$ is defined as zero when $x^r_i=c^r_l$, and so is $d_a^l(.)$. 
\begin{figure}[t]
\centering
\includegraphics[width=.8\columnwidth]{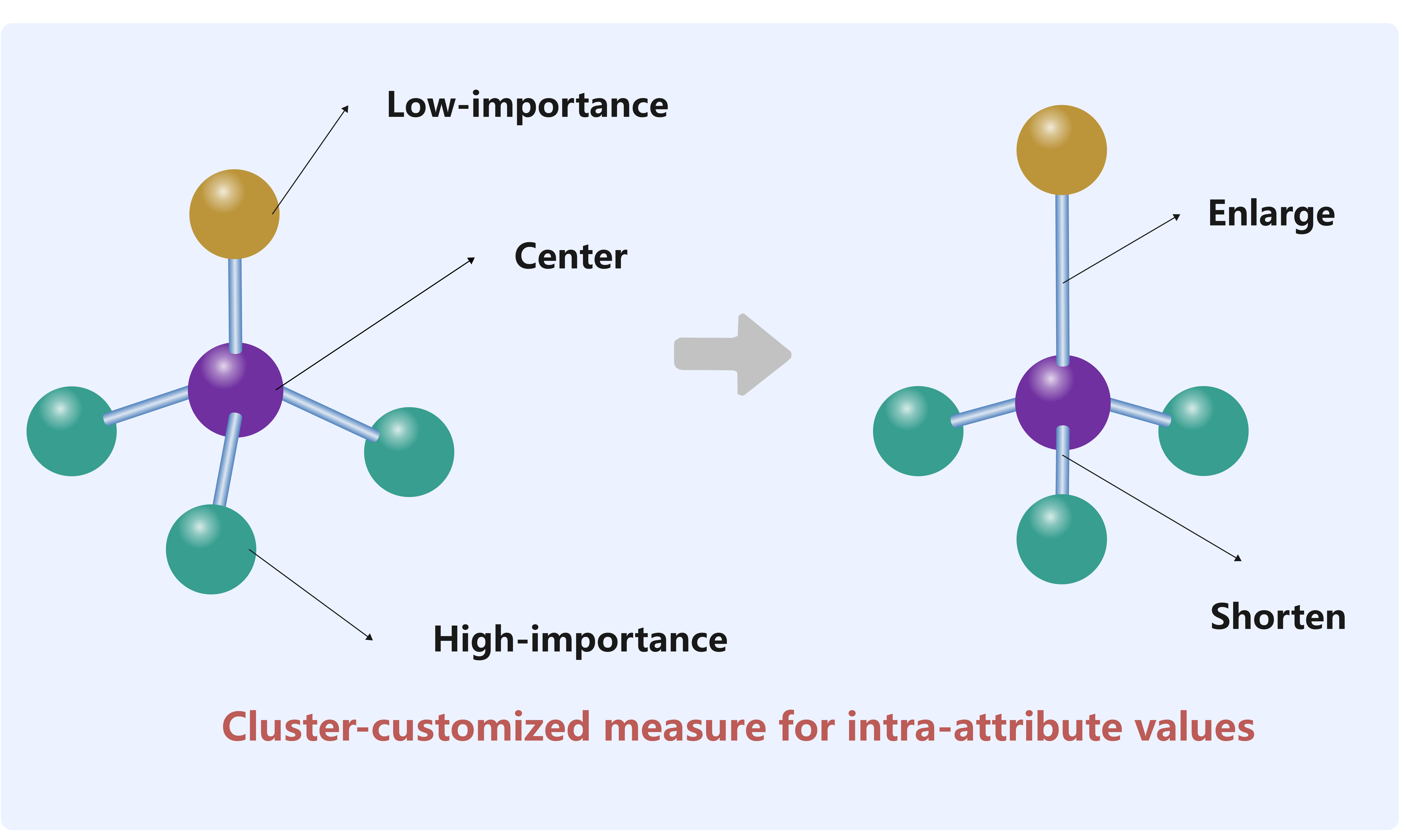} 
\caption{Framework of attribute value distance measurement}
\label{compare}
\end{figure}

Furthermore, as shown in Fig~\ref{compare}, the CVD is proposed to cluster-customized measure the distance between attribute values, which is defined as:
\begin{equation}
   d_a^l(o_s^r, o_p^r) = \gamma^l(o_s^r) + \gamma^l(o_p^r),
    \label{eq: dis}
\end{equation}
where the $o_s^r$ is used to represent $o^r_j$ and $o^r_t$ for generality reasons, and it is called the rival attribute value. $\gamma^l(.)$ is the rival factor for attribute values. It is designed to construct the CVD based on the CVI of both cluster center and categorical data for reasonable measurement.
\begin{remark}
    As Fig~\ref{compare} shows, $d_m(.)$ is a cluster-customized measure that can adaptively change in different clusters. CVD defines that the data with high CVI is closer to the cluster center, as it represents this cluster. Otherwise, it should be far away from the cluster center. This can be regarded as a rival process between different $o^r_s$ and $o^r_p$ in each cluster. Thus, we define a rival factor as a bridge between the CVD and CVI to achieve this rival process for measuring the distance of attribute values.
\end{remark}
Specifically, we define the rival factor as:
\begin{equation}
 \gamma^l(o^r_z) =\left\{
\begin{array}{l l}
CVI^l(o_p^r), & o^r_z \in o^{r}_p \\[1ex]
\frac{1}{CVI^l(o_s^r)}. & o^r_z \in o^r_s
\end{array} \right.
\label{eq:intra2}
\end{equation}
Since the cluster center's attribute value ($o^r_p$) is generally of high importance, its rival factor should contribute greatly to the distance based on CVI, and it is the basis of the rival process. Moreover, the rival factor for the rival attribute value ($o^r_s$) is a reciprocal form of the cluster center's rival factor $\gamma^l(o^r_p)$. The intuition is that the rival factor of the rival attribute value will enlarge the distance when the rival attribute value's CVI is low. The CVI is computed by:
\begin{equation}
   CVI^l(o_s^r) = \frac{C^l(o_s^r)}{\displaystyle\max_{\substack{\\1 \leq f \leq v^r}}C(o_f^r)},
    \label{eq: intra-d}
\end{equation}
\begin{algorithm}[tb]
\caption{Proposed CADM clustering algorithm}
\label{alg:algorithm}
\textbf{Input}: Dataset \textbf{S}, number \textbf{k} of clusters\\
\textbf{Output}: cluster label \textbf{L}
\begin{algorithmic}[1] 
\WHILE{$C^t \neq C^{t-1}$ }
\STATE Calculate the distance between attribute values based on the CVD by Eq.(\ref{eq: dis}).
\STATE Utilizing Eq.(\ref{eq: self}) to obtain the attribute importance to further constrain the distance.
\STATE Gaining final distance measurement based on Eq.(\ref{eq: logic}).
\STATE Update \textbf{D}, \textbf{C} and \textbf{L}.
\ENDWHILE
\STATE \textbf{return} \textbf{L}
\end{algorithmic}
\end{algorithm}
which offers the relative importance of the intra-attribute value. The $C^l(.)$ provides the count of one attribute value in a specific cluster. Differently, $C(.)$ provides the total counts of one attribute value in the whole dataset $S$. Thus, CVI is the frequency count of the rival attribute value. In this case, the CVI will adaptively update in each iteration and cluster.
Moreover, the importance of different attribute categories varies, and they are different in other clusters. Thus, CAI is defined to explicitly weigh the contributions of attribute categories in forming distances, which is computed by:
\begin{equation}
     CAI^l(A^r) = \frac{\displaystyle\max_{\substack{\\1 \leq s \leq v^r}} C^l(o^r_s)}{n},
    \label{importance}
\end{equation}
Thus, CAI can be leveraged to define attribute importance, which is computed by: \begin{equation}
    d_I(A^r) = CAI^l(A^r)^2.
    \label{eq: self}
\end{equation} 
\begin{remark}
CAI is calculated as a cohesion factor considering the consistency of the count of the possible attribute values in specific attribute $A^r$. Thus, the higher the maximum count of possible attribute values, the more consistent the attribute, and the more weight will be added to the distance calculated in this attribute.
\end{remark}

The whole algorithm process is shown in \textbf{Algorithm \ref{alg:algorithm}}. It adopts the framework of K-modes clustering that iteratively updates cluster center $C$, distance matrices $D$, and cluster labels $L$ in each time step $t$ until convergence.

\begin{table*}[t]
\centering
\Huge
\caption{Experiments with competitive distance metric in categorical, ordinal, and nominal datasets. "$-$" indicates that the algorithm is inapplicable or has not converged in one dataset.}
  \resizebox{.9\linewidth}{!}{
    \begin{tabular}
    {c c c || c  | c | c | c| c| c| c |c |c| c}
          \toprule[1.8pt]
          \multicolumn{3}{c||}{Dataset Statistics}  &  HDM~\cite{hdm} & GSM~\cite{gsm} & LSM~\cite{LSM} & CBDM~\cite{ahmad} & EBDM~\cite{unified1} & UDM~\cite{udm} & HARR~\cite{HARR} & COF~\cite{COF} & QGRL~\cite{qgrl} &\textbf{CADM} \\
          \cline{1-3}
           Abbrev. & $\#$Instance & $k$ & Baseline & Baseline & Baseline & 2012 & 2020 & 2022 & 2025 & 2024 & 2024 &Ours \\
 
        \midrule[1pt]
         NS & 12960 & 4  & $0.375\pm 0.04$ & $0.356\pm 0.03$ & $0.375\pm 0.03$ & $-$ & $0.400\pm 0.02$ & $\underline{0.411\pm0.03}$ & $0.407 \pm0.02$ &$0.362\pm0.09$ & $0.395\pm0.02$
        &$\mathbf{0.429\pm0.03}$  \\
        PR & 123& 12 & $0.410\pm 0.04$ & $0.393\pm 0.04$ & $0.396\pm 0.04$ & $0.399\pm 0.03$ & $0.361\pm 0.04$ & $0.412\pm0.03$ &$0.431\pm 0.06$ &$0.429\pm0.03$ & $\mathbf{0.678\pm 0.02}$
        &$\underline{0.433\pm0.05}$ \\
        HA&132&3 & $0.389\pm 0.02$ & $0.398\pm 0.04$ & $0.392\pm 0.04$ & $0.383\pm 0.04$ & $0.407\pm 0.03$ & $0.446\pm0.04$ & $0.447\pm0.03$ &$\underline{0.453\pm0.02}$&$0.362\pm0.02$ & $\mathbf{0.471\pm0.03}$ \\
        LY& 148 & 4 & $0.459\pm 0.05$ & $0.451\pm 0.04$ & $0.459\pm 0.05$ & $0.489\pm 0.05$ & $0.450\pm 0.03$ & $\underline{0.494\pm0.03}$ &$0.453\pm 0.04$ &$0.488\pm 0.12$ & $0.462\pm 0.03$ &
        $\mathbf{0.507\pm 0.04}$ \\
        SM&61069 & 2 &$0.506 \pm 0.01$&  $0.508 \pm 0.01$ &$\underline{0.530 \pm 0.01}$ &$-$& $0.520 \pm 0.02$ & $0.521 \pm 0.01$ & $0.516 \pm 0.02$ & $0.504 \pm 0.02$ & $-$
        &$\mathbf{0.550 \pm 0.03}$\\
        C4&67577&3 &$ 0.371 \pm 0.03$&$ 0.373 \pm 0.03$&$ 0.358 \pm 0.01$ & $-$ & $0.356 \pm 0.04$ & $0.378 \pm 0.02$ & $0.383 \pm 0.03$ &$\mathbf{0.431 \pm 0.03}$ &$-$&$\underline{0.411 \pm 0.03}$\\
        VT & 435 & 2 & $0.874\pm 0.01$ & $0.534\pm 0.01$ & $0.534\pm 0.00$ & $0.806\pm 0.01$ & $0.853\pm 0.00$ & $0.872\pm 0.00$ & $0.873\pm 0.01$&$0.875\pm 0.01$
        &$\mathbf{0.884\pm 0.02}$
        &$\underline{0.880\pm 0.00}$  \\
       LS & 24 & 3 &$0.375\pm 0.02$ &$0.502\pm 0.01$ &  $\underline{0.595\pm 0.03}$ & $0.515\pm 0.01$ & $0.508\pm 0.02$ & $0.550\pm 0.03$ &  $0.501\pm 0.03$& $0.563\pm 0.08$& $-$&
       $\mathbf{0.608\pm 0.03}$  \\
        PE & 101 & 7 & $0.485\pm 0.03$ & $0.515\pm 0.03$ & $0.419\pm 0.02$  &$0.409\pm 0.02$ &$\underline{0.610\pm 0.03}$ & $0.609\pm 0.02$ & $0.545\pm 0.03$ & $0.561\pm 0.04$& $0.557\pm 0.03$& $\mathbf{0.615\pm 0.04}$  \\
        LE& 1000 & 5 &  $0.269\pm 0.04$ & $0.298\pm 0.03$ & $0.303\pm 0.03$ & $0.306\pm 0.02$ & $0.369\pm 0.02$ &$\underline{0.372\pm 0.03}$ & $0.345\pm 0.04$ &$0.319\pm 0.06$ & $0.337 \pm 0.02$ &$\mathbf{0.373\pm 0.02}$  \\
        AA & 104 & 2 & $0.577\pm 0.01$ & $0.510\pm 0.02$ & $0.576\pm 0.02$ & $-$ & $0.601\pm 0.03$ & $0.567\pm0.03$ & $0.560 \pm0.02$ &$0.559\pm0.04$ &$\underline{0.636\pm0.01}$ &$\mathbf{0.661\pm0.03}$  \\
        HF& 299 & 2 & $0.599\pm 0.02$ & $0.679\pm 0.01$ & $0.602\pm 0.02$ & $-$ & $0.625\pm 0.03$ & $0.600\pm0.02$ &$0.704\pm 0.03$ &$0.692\pm0.02$ & $\underline{0.713\pm0.03}$ &$\mathbf{0.736\pm0.03}$ \\
        HD & 297 & 5 & $0.351\pm 0.02$ & $0.358\pm 0.04$ & $0.391\pm 0.04$ & $-$ & $0.360\pm 0.03$ & $0.377\pm0.04$ & $0.417\pm0.03$ &$0.403\pm0.04$&$\underline{0.432\pm0.02}$
        &$\mathbf{0.471\pm0.03}$ \\
        MM & 824 & 2 & $0.818\pm 0.00$ & $0.820\pm 0.00$ & $0.831\pm 0.00$ & $0.828\pm 0.00$ & $0.807\pm 0.00$ &$\mathbf{0.837\pm0.00}$ &$0.818\pm 0.00$& $0.826\pm 0.01$ &  $0.830\pm0.00$ &$\underline{0.832\pm0.00}$  \\
        \midrule[1pt]
        &&Rank: & 7.1 & 7.7 & 6.6 & 7.0 & 6.1 & 3.9 & 4.9 &4.6& \underline{3.0} & $\mathbf{1.3}$ \\
        \bottomrule[1.8pt]
        
    \end{tabular}
    }
  \label{tab:categorical}
\end{table*}
\section{Experiment}
\label{sec: experiment}
\textbf{Nine Counterparts:} including three classical (i.e.,  HDM \cite{hdm}, GSM~\cite{gsm}, LSM \cite{LSM}), two context-based (CBDM~\cite{ahmad}, EBDM~\cite{unified1}), and four SOTA (i.e.,UDM~\cite{udm}, HARR~\cite{HARR}, COF~\cite{COF}, QGRL~\cite{qgrl}) clustering algorithm are chosen. Especially, QGRL is a deep learning based algorithm, while others are unsupervised learning. Set the cluster number $k$ as the real class number of the data. We ran ten times for each experiment and used the average value in the report table.

\textbf{Fourteen Datasets:} are collected from \cite{weka} and \cite{uci} shown in Table~\ref{tab:categorical}, including 4 mixed (i.e., AA, HF, HD, MM), 5 categorical (i.e., NS, PR, HA, LY, SM), 3 ordinal (i.e., C4, LE, LS), and 2 nominal datasets (i.e., VT, PE). 

\textbf{Validity Indices}: Clustering Accuracy (CA) \cite{ca} is selected for evaluating the clustering performance. Larger values
indicate better clustering performance. 

\begin{figure}[t]
\centering
\subfigure[]{
    \includegraphics[width=0.5\columnwidth]{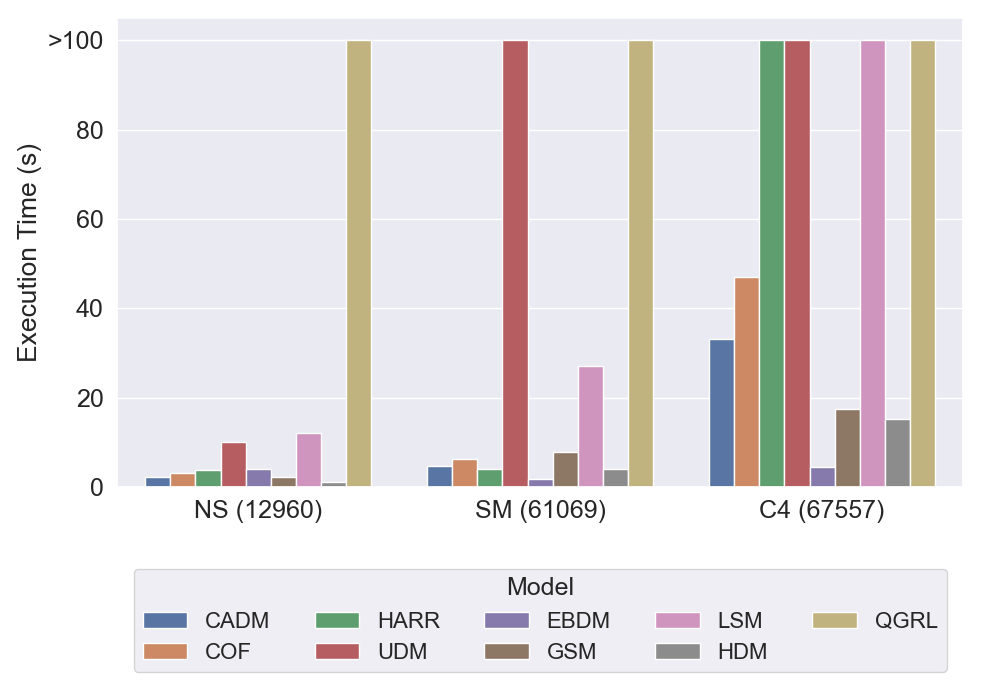}
}
\hspace{-0.35cm}
\subfigure[]{
    \includegraphics[width=0.47\columnwidth]{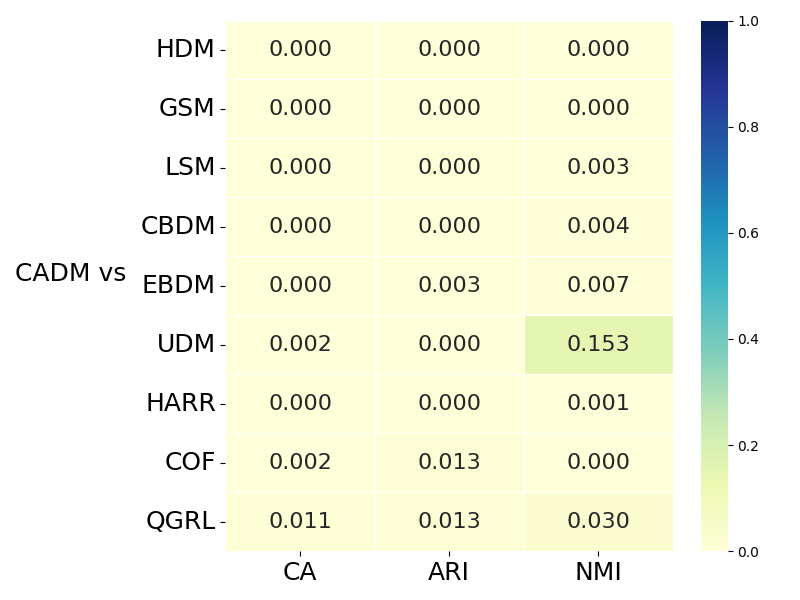}
}

\subfigure[]{
\vspace{-1pt}
    \includegraphics[width=0.47\columnwidth]{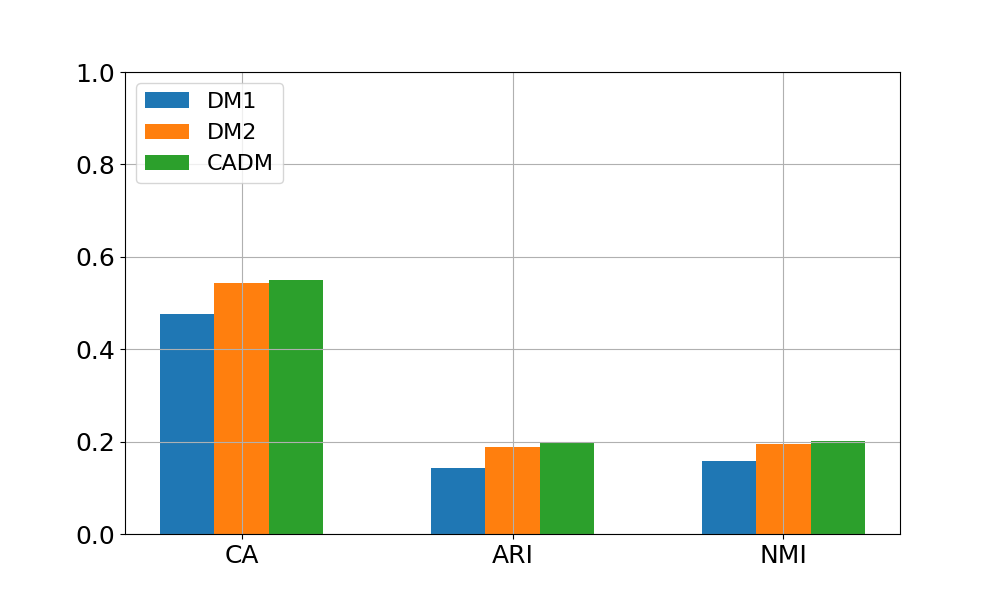}
}
\hspace{-0.3cm}
\vspace{-0.3cm}
\subfigure[]{
    \includegraphics[width=0.47\columnwidth]{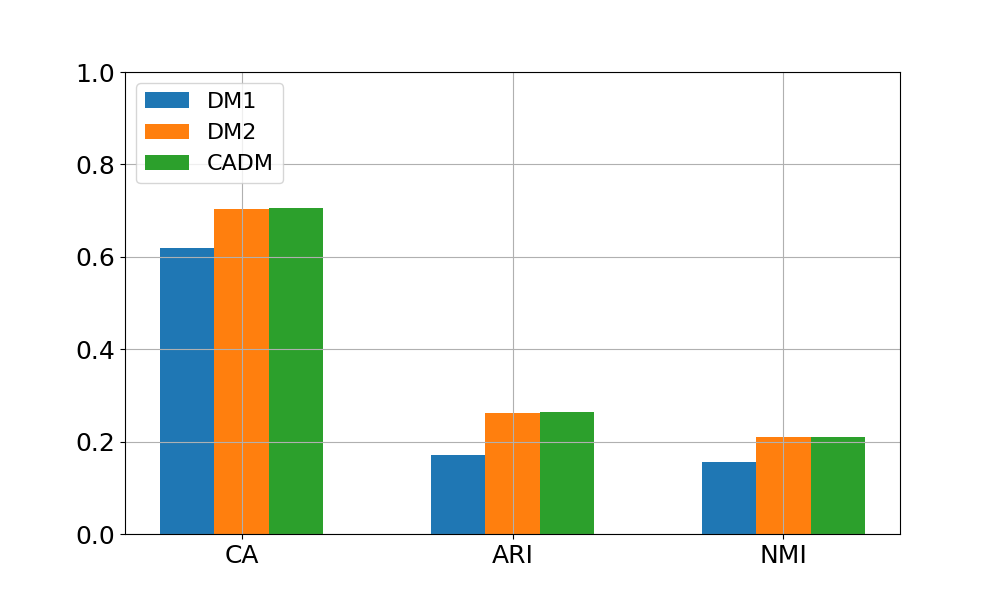}
}

    \caption{(a) efficiency test on three large datasets. (b) Wilcoxon signed rank test in fourteen datasets. (c) and (d) demonstrate the ablation study in the categorical and mixed datasets.}
        \label{subfig:time_three}
\end{figure}

\textbf{Comparative Results:} The bold value means the best performance in a dataset, and the underline value means the second-best performance in one dataset. As Table~\ref{tab:categorical} shows, the proposed CADM outperforms nine counterparts with an average rank $1.3$, indicating its superiority in categorical and mixed data clustering. On categorical datasets (i.e., NS, LY, SM), the advantage of CADM is extremely obvious, indicating that the proposed cluster-personalized metric can provide more accurate distance information for each cluster. The superiority of CADM on the mixed dataset (i.e., AA, HF, HD) is also tremendous, which illustrates its significant universality in heterogeneous datasets. Moreover, Fig~\ref{subfig:time_three} (b) shows the results of the Wilcoxon signed rank test between our CADM and the other nine methods in fourteen datasets, which indicates CADM has a significant superiority over other methods, achieving a 95\% confidence level.

\textbf{Efficiency Evaluation :} We select three large datasets (i.e., NS, SM, C4) to examine model efficiency. Based on the Fig~\ref{subfig:time_three} (a), CADM outperforms the four latest SOTA models. Although three baselines are faster, their clustering performance is extremely lower than CADM in fourteen datasets.

\textbf{Ablation studies:} In ablation studies, DM1 is a simple distance measurement leveraging order information. DM2 adds the CVD, and CADM adds the CAI. The results in Fig~\ref{subfig:time_three} (c) and (d) illustrate the effectiveness of the proposed cluster-customized meteic framework. Specifically, it is obvious that CVD drastically improves the performance, indicating the benefits of the cluster-customized framework, and CAI also effectively adjusts the final measurement. More comparative results (in other indicators), complexity analysis, and proofs can be found in \href{https://anonymous.4open.science/r/CADM-47D8/Appendix.pdf}{online appendix.} 

\section{Concluding Remarks}
This paper proposes a novel cluster-customized adaptive distance metric for categorical data clustering. It is a unified distance metric for categorical data, which is applicable to both nominal and ordinal data. Specifically, Cluster-customized attribute value distance measurement is defined considering the competitive cluster-customized strategy to address the concern of the distance difference between two attribute values in various clusters. Besides, the importance of the attribute has been proposed to weigh the contributions of different attributes in forming distance, making the distance measurement more reasonable. Experiments have shown CADM's superiority in categorical data clustering. Moreover, it is efficient without any pre-set parameters, and its mechanisms have high interpretability, indicating its significant potential. 

\section{Acknowledgment}
This work was supported in part by the HKBU Seed Funding for Collaborative Research Grants under grant: RC-SFCRG/23-24/R2/SCI/10, and the Guangdong and Hong Kong Universities “1+1+1” Cross-Campus Research Collaboration Scheme with the grant: 2025A0505000004.

\begin{spacing}{0.1} 
    \setlength{\itemsep}{-3mm} 
    \bibliographystyle{IEEEbib} 
    \bibliography{cadm} 
\end{spacing}
\end{document}